\documentclass{article}

\usepackage{PRIMEarxiv}

\usepackage[utf8]{inputenc} 
\usepackage[T1]{fontenc}    
\usepackage{hyperref}       
\usepackage{booktabs}       
\usepackage{amsfonts}       
\usepackage{nicefrac}       
\usepackage{microtype}      
\usepackage{lipsum}
\usepackage{fancyhdr}       
\usepackage{graphicx}       
\graphicspath{{media/}}     

\usepackage{amssymb}

\usepackage[acronym]{glossaries}
\usepackage{glossary-inline}
\usepackage{url}
\usepackage{amsmath}
\usepackage{subcaption}
\usepackage{lscape} 
\usepackage{algorithm}
\usepackage{algorithmic}
\usepackage{multirow}
\usepackage[official]{eurosym}
\usepackage{makecell}
\usepackage{xcolor}
\usepackage[labelfont=bf]{caption}
\usepackage{setspace}
\usepackage{stfloats}
\usepackage{bigfoot}
\usepackage{multirow}

\makeglossaries
\newacronym{rl}{RL}{Reinforcement Learning}
\newacronym{drl}{DRL}{Deep Reinforcement Learning}
\newacronym{ppo}{PPO}{Proximal Policy Optimisation}
\newacronym{re}{RE}{Renewable Energy}
\newacronym{ml}{ML}{Machine Learning}
\newacronym{pbt}{PBT}{Population Based Training}
\newacronym{mdp}{MDP}{Markov Decision Process}
\newacronym{sgd}{SGD}{Stochastic Gradient Decent}
\newacronym{tso}{TSO}{Transmission System Operators}
\newacronym{opf}{OPF}{Optimal Power Flow}
\newacronym{l2rpn}{L2RPN}{Learning to Run a Power Network}
\newacronym{gnn}{GNN}{Graph Neural Network}
\newacronym{smaac}{SMAAC}{Semi Markov Afterstate Actor Critic}
\newacronym{ddqn}{DDQN}{Dueling Deep Q Network}
\newacronym{bohb}{BOHB}{Bayesian Optimisation Hyperband}

\newacronym{t_o}{$Tutor_{original}$}{Original Tutor Agent}
\newacronym{t_n1}{$Tutor_{N-1}$}{Tutor Agent with N-1 strategy}
\newacronym{t_n1_topo}{$Tutor_{N-1, Topo}$}{Tutor Agent with N-1 strategy and the topology reversion}
\newacronym{senior}{$Senior_{N-1,Topo}$}{Senior Agent with N-1 action and topology reversion}
\newacronym{d_n}{$Do\_Nothing$}{Do-nothing Agent}
\newacronym{exp}{$Expert$}{Expert Agent}

\DeclareMathOperator*{\argmin}{\arg\!\min}

\title{Managing power grids through topology actions: A comparative study between advanced rule-based and reinforcement learning agents
}

\author{
  Malte Lehna \thanks{\textbf{Corresponding Author}: \texttt{malte.lehna@iee.fraunhofer.de}, \url{https://orcid.org/0000-0003-0621-1442}} \\
  Fraunhofer IEE \\
  Kassel \\
  Germany\\
  \\
   \And
  Jan Viebahn \\
  TenneT TSO BV \\
  Arnhem \\
  Netherlands\\
  \And
  Christoph Scholz\thanks{\url{https://orcid.org/0000-0002-8719-8261}} \\
  Fraunhofer IEE \\
  Kassel \\
  Germany\\
  \And
  Antoine Marot \\
  AI Lab, Reseau de Transport d’Electricite (RTE)\\
  Versigny\\
  France\\
  \And
  Sven Tomforde\thanks{\url{https://orcid.org/0000-0002-5825-8915}} \\
  Kiel University: Intelligent Systems \\
  Kiel \\
  Germany\\
}

\begin{document}
\maketitle

\begin{abstract}
The operation of electricity grids has become increasingly complex due to the current upheaval and the increase in renewable energy production. As a consequence, active grid management is reaching its limits with conventional approaches. In the context of the \gls{l2rpn} challenge, it has been shown that \gls{rl} is an efficient and reliable approach with considerable potential for automatic grid operation. In this article, we analyse the submitted agent from Binbinchen and provide novel strategies to improve the agent, both for the \gls{rl} and the rule-based approach. The main improvement is a N-1 strategy, where we consider topology actions that keep the grid stable, even if one line is disconnected. More, we also propose a topology reversion to the original grid, which proved to be beneficial. The improvements are tested against reference approaches on the challenge test sets and are able to increase the performance of the rule-based agent by 27\%. In direct comparison between rule-based and \gls{rl} agent we find similar performance. However, the \gls{rl} agent has a clear computational advantage. We also analyse the behaviour in an exemplary case in more detail to provide additional insights. Here, we observe that through the N-1 strategy, the actions of the agents become more diversified. 

\end{abstract}


\section{Introduction}
\label{sec:introduction}
\printglossary[type=\acronymtype,style=inline]
\subsection{Overview}
\label{ssec:motivation}
With the growth of renewable energies in the electricity mix and their volatile behaviour, the complexity of operating an electricity grid is constantly increasing for \gls{tso} \cite{marot2022perspectives}. As a result, it is not only necessary to plan production capacities and predict the demand correctly, but also consider new options to manage grids in times of instabilities. Topological changes at substation level are an option that is gaining increasing attention as this is an existing cheap but underutilised flexibility. Their usage in controlling the stability of electricity grids has been discussed by several researchers over the last decades \cite{bacher1986network}. 
However, changes in topology also have a major drawback, namely that their optimisation requires a large amount of computational resources \cite{viebahn22}. In the past, \gls{opf} optimisation has often been used, but researchers have noted potential difficulties with the increase in renewable energy and smart grids  \cite{capitanescu2016critical} as well as  criticised its inability to cope with large, non-linear combinatorial action spaces \cite{marot2020l2rpn}. 
To solve this problem, the French \gls{tso} RTE proposed with the \glsfirst{l2rpn} a \glsfirst{rl} challenge that aims to join forces between the grid experts and the \gls{ml} community.\footnote{\gls{l2rpn} challenges: \url{https://l2rpn.chalearn.org/} (last access 20/03/2023).} In the challenge, \cite{marot2020learning} provided a \gls{rl} environment with the Grid2Op package \cite{grid2op}, which allowed research on real data, while allowing the training of different algorithms on electricity grids.\footnote{Grid2Op: \url{https://github.com/rte-france/Grid2Op} (last access 20/03/2023).} Grid2Op has become one of the leading frameworks for \gls{rl} grid control in scenario-based simulations. We therefore conduct our research in this environment in order to provide comparable results. 

\subsection{Research Contribution}
\label{ssec:res_contr}
While the first contributions to the automation of grid control have been made \cite{marot2021learning}, it is not yet clear how large the impact of the \gls{rl} solution will be. Is the \gls{rl} approach solely responsible, or can heuristic and rule-based approaches achieve a similar solution if their behaviour is sophisticated enough?\\ 
In our work, we contribute to this discussion by providing a systematic analysis between a rule-based agent and a \gls{rl} agent. For this, we have chosen the framework of Binbinchen \cite{binbinchen} from the 2020 \gls{l2rpn} robustness challenge, as it includes both a rule-based and a \gls{rl} agent. As part of our work, we extend the rule-based approach with two significant improvements. First, we propose an N-1 strategy to ensure more robust topology actions of the rule base agent.
This N-1 strategy favours topology actions that ensure a stable grid, even if one line of the grid is disconnected. Second, we encourage the agent to revert to the original topology when the state of the grid is relatively stable. We analysed the effect of these improvements in an experiment, where we ran the scenarios with 30 different seeds to ensure an significant evaluation with less randomness. Our experiments show that with our improvement we were able to increase the performance of the rule-based agent by $27\%$ in comparison to the original Binbinchen agent. Furthermore, we are able to show that our advanced rule-based approach is able to achieve similar performance to the \gls{rl} agent. However, when analysing the computational cost, the \gls{rl} agent is still advantageous. In order to use the agents on all Grid2Op environments, we reworked the original code and published it as an open-source package on Github.\footnote{CurriculumAgent: \url{https://github.com/FraunhoferIEE/CurriculumAgent} (last access 20/03/2023).} 
Overall, our contributions can be summarised as follows: 
\begin{enumerate}
    \item We analysed the solution of \cite{binbinchen} and transferred its contents from a rapid prototyping state to a Python package. With various algorithmic changes, we ensure the usability for other grids without hard-coded implications based solely on the robustness challenge. This allows the community to include methods from the CurriculumAgent and compare their agents against our baseline.
    \item We introduce two important improvements (N-1 strategy and topology reversion) to the rule-based agent that dramatically increase its performance by up to $27\%$ in comparison to the original agent. 
    \item We apply the above modifications to the \gls{rl} agent and benchmark its performance with the rule-based method. 
    \item We propose a test framework with multiple seeds to ensure significant comparable results.
\end{enumerate}
The remainder of this article is structured as follows. In Section~\ref{sec:rel_work}, we provide the related research and afterwards outline the structure of the Grid2Op environment in Section \ref{sec:grid2op}. In Section~\ref{sec:method}, we then introduce the \textit{Teacher-Tutor-Junior-Senior} framework, followed by our novel improvements in Section~\ref{sec:improvements}.
In Section~\ref{sec:framework}, we present the experimental setup of this article and afterwards, in Section~\ref{sec:results}, we discuss the results. Lastly, we provide a conclusion in Section~\ref{sec:conclusion}.

\section{Related Work}
\label{sec:rel_work}
In recent years, there has been an increasing interest in the usage of \glsfirst{rl} for the control of power systems. 
One early paper was published by \cite{ernst2004power} in 2004, where the authors proposed to use the \gls{rl} approach for both offline and online applications in the context of power systems. However, the use of \gls{rl} was still constrained due to the computational limitations of that time. Everything changed with the breakthrough of \gls{drl}, first introduced by \cite{mnih2015human,mnih2016asynchronous}. 
Their work set off a wave of research, with ground breaking results in various fields \cite{lee2020learning,schrittwieser2020mastering,berner2019dota}. The idea behind \gls{drl} is that the policies and/or value functions of the algorithms are not computed directly, but instead are approximated through a deep neural network. This allows the learning of advanced strategies, even for complex problems.\footnote{Note that for the sake of simplicity, we henceforth refer to \gls{drl} as \gls{rl} in this paper.} 

Following these breakthroughs, researchers began addressing grid control problems with \gls{rl}, especially driven by the introduction of the \gls{l2rpn} challenge \cite{marot2020l2rpn,kelly2020reinforcement, 9494879, marot2021learning}.
The first successful \gls{rl} solution was presented by participants in the
\gls{l2rpn} challenge 2019 \cite{lan2020ai}, who used the \gls{ddqn} agent as their \gls{rl} approach  and were able to win the competition. The authors in \cite{lan2020ai} also introduced for a first time an imitation learning strategy and combined it with an guided exploration approach. In the following L2RPN WCCI 2020, other participants implemented a \gls{smaac} algorithm\cite{yoon2021winning}, which focused on the final topology instead of the switching of individual actions at the substation level to learn the different topology actions. By applying a \gls{gnn} to the observation space, \cite{yoon2021winning} further transformed the original grid into an afterstate representation. With their approach, \cite{yoon2021winning} outperformed all other candidates and became the first winner of the challenge. 
Following the first challenge, \cite{marot2020l2rpn, marot2021learning} increased the difficulty for \gls{rl} agents in the \gls{l2rpn} NeuriPS 2020 challenge, by introducing a robustness track with an adversarial agent and an adaptability track with  an increasing share of renewable energy injections \cite{marot2020l2rpn}. Interestingly, neither track could be solved by the previous agents, instead new solutions were required.
One submission was published in \cite{zhou2021action}, where the authors used an evolutionary \gls{rl} approach in order to win both challenge tracks. They introduced a planing algorithm to actively search through the available actions provided by the policy. Afterwards, the planing algorithm was optimised through an evolution strategy. Within the black box optimisation method, Gaussian white noise was added to ensure an adequate exploration of the policies. Apart from to topology actions, the authors also included line reconnecting actions and some hand picked redispatch actions\cite{zhou2021action}.
The second best performance of the robustness track was the agent by Binbinchen \cite{binbinchen}, already mentioned in Section \ref{ssec:res_contr}. They proposed a framework, where a \gls{rl} agent learned though imitating a rule-based agent, ensuring a better handling of the large action space. From a methodological perspective, their agent was based on the \gls{ppo} algorithm by \cite{schulman2017proximal}. Due to their performance and their straightforward framework, the approach of \cite{binbinchen} is the foundation of our work and we describe their framework in more detail in Section~\ref{sec:agent}. \\
One notable approach that was not part of the challenge, but instead was published recently, was the paper \cite{chauhan2022powrl}. The researchers combined a \gls{rl} algorithm with an heuristic approach, sharing several similarities to the agent of \cite{binbinchen} and our work. With their combination, they were able to achieve better results than any previous agents on the legacy data set of the 2020 \gls{l2rpn} robustness challenge, while at the same time reducing the overall run-time of their agent.  
Recently, another agent was proposed in \cite{dorfer2022power}, based on the AlphaZero \cite{silver2017mastering}. Similar to AlphaZero, the researchers of \cite{dorfer2022power} used the Monte Carlo Tree Search to find appropriate actions in the grid. 
However, some specifications were changed to the use-case of the grid management. These changes were an early stopping criteria to tackle the expensive computational costs and a heuristic model. The latter was used instead of a neural network to describe the agent's value function, further reducing the training time \cite{dorfer2022power}. With their approach, they were not only able to win the \gls{l2rpn} WCCI challenge of 2022, but also achieved with only topology actions the same results as agents with redispatching actions. This is indicates the importance of topology actions for future research.\\
Finally, given that we are also interested in rule-based approaches, we would like to emphasise the expert agent of \cite{marot2018expert}. In their work, the researchers build an agent that searched a priori for remedies
by identifying local power paths to manage the electricity grid. In their agent, they include expert knowledge to rank the topology actions and chose the correct action in case the grid is unstable. Considering that the agent is also a good starting point for our rule-based approach, we include the agent as baseline in our work. 
 
\begin{figure*}[!t]
    \centering
    \caption{Visualisation of an exemplary topology action, adapted from the paper of \cite{marot2020learning}. The original grid shows an overload of the right line (in red) at time step $t$, due to an high demand from both load sinks. By executing a topology action, i.e., splitting the load flow of the substation into two separate notes, the bottom-right substation can divert the power and the grid returns to a more stable state without an overflow. }
    \includegraphics[trim={0cm 0cm 0cm 0.0cm},clip, width=0.8\linewidth]{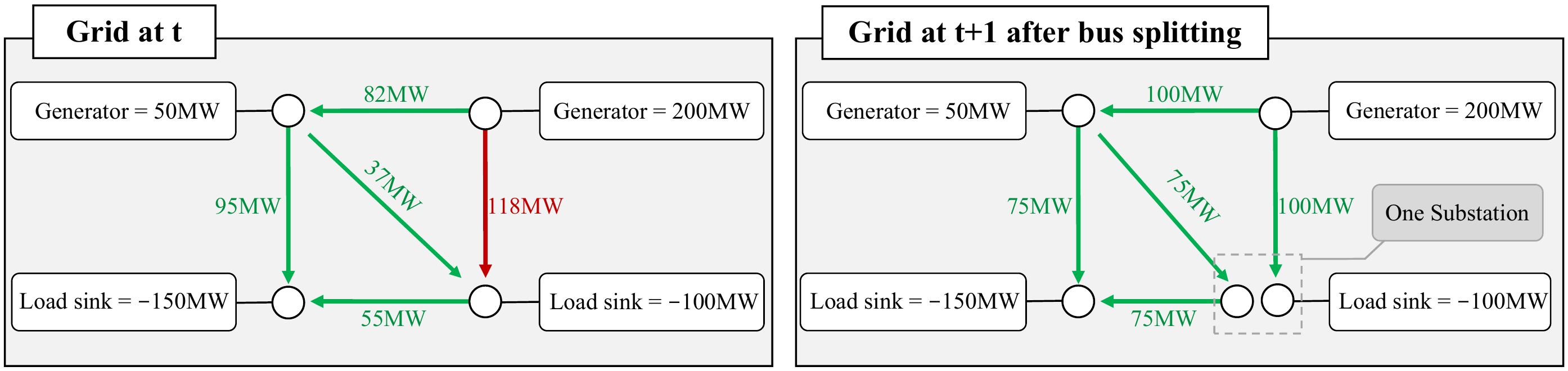}
    \label{fig:topo_act}
\end{figure*}
\section{The Grid2Op Environment}
\label{sec:grid2op}
\begin{figure*}[!t]
    \centering
    \caption{The electricity grid of the robustness track, based on a subset of the IEEE118 grid. In the grid, a total of 35 substations exist that are interconnected with power lines. The grid has both generators and load sinks in different parts of the grid. As original state, all power lines are connected to bus 1 on the substations. Through topology actions, these can be changed to bus 2.  The figure was created with the internal plot method of Grid2Op.}
    \includegraphics[trim={0cm 0.5cm 0cm 0.5cm},clip, width=0.8\linewidth]{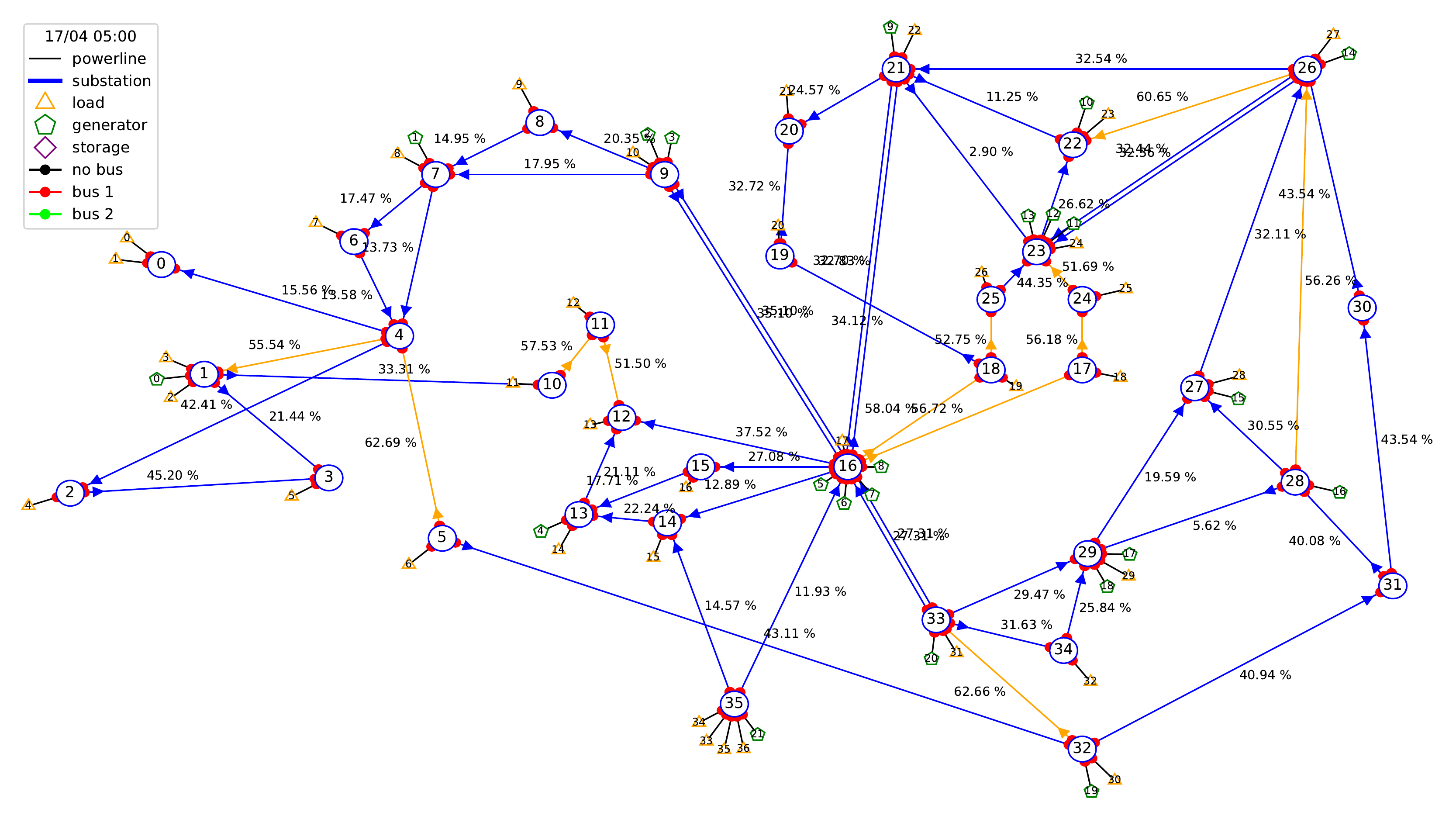}
    \label{fig:grid}
\end{figure*}

The Grid2Op environment is a Python package created by RTE to provide a platform for the development of \gls{rl} agents. As outlined in \cite{marot2020learning,kelly2020reinforcement}, the package enables training and the evaluation of agents in the \gls{rl} Gym framework \cite{brockman2016openai} on known IEEE synthetic power grids, e.g, the IEEE14 or IEEE118 grids.\footnote{Gym-framework: \url{https://www.gymlibrary.dev/}  (last access 20/03/2023).} To correctly depict an \gls{mdp}, the environment of Grid2Op consists of an observation space, an action space and provides multiple reward function. The action space is divided into three action types. The first type of action is the line action $a^{(line)} \in \ \mathcal{A}^{(line)}$ that connects and disconnects lines between two substations.
The second type of action is a topological action $a^{(topo)}$ from the set of all possible topology actions $a^{(topo)} \in \ \mathcal{A}^{(topo)}$. The topology action changes the node configuration on a substation level. This is visualised in Figure \ref{fig:topo_act}, where the node splitting is demonstrated on a simple example case, based on \cite{marot2020learning}. Note that the complexity increases with more lines connected to a substation \cite{9494879}. As a third type of action, Grid2Op offers the possibility to change the power injection through redispatch actions with $a^{(redisp)} \in \mathcal{A}^{(redisp)}$, by adjusting the production of the generators in the grid. While the action spaces $\mathcal{A}^{(line)}$ and $\mathcal{A}^{(topo)}$ are each discrete action spaces, $\mathcal{A}^{(redisp)}$ is considered continuous.
With respect to the observation space, Grid2Op provides various information on the grid that range from topology information and power flows to line capacity and time variables as well as the injection of the generators and demand of the consumers. While all the information is vital in the training of the agents, especially the capacities of the lines are used to measure the stability of the grid.
In terms of Grid2Op, the capacity of a line is defined as the observed current flow of the line, divided through its thermal limit. 
We denote the capacity of a line $l$ as $\rho_{l,t} \in \mathbb{R}^+$, with $l={1,\ldots,L}$ from the set of all lines $l \in \mathcal{L}$. We further recorded for each time step $t$ the maximum capacity across all lines as $\rho_{max,t}=\max\limits_{l=1,\ldots,L} (\rho_{l,t})$.
The grid is considered unstable, if at least one line is above its capacity of 100\%. In addition to the actual observation, Grid2Op offers the possibility to simulate the effect of an action on the grid. This simulation method is not fully accurate, as it relies on forecasted values. We denote this simulation of the line capacity as $\hat{\rho}_{l,t+1}$ and the maximum of the simulated value as 
$\hat{\rho}_{max,t+1}=\max\limits_{l=1,\ldots,L} (\hat{\rho}_{l,t+1})$. \\ 
In our work, the robustness track of the 2020 \gls{l2rpn} robustness track was chosen as benchmark. Here, the grid consisted of one out of three regions from the IEEE118 grid, shown in Figure~\ref{fig:grid}. This corresponds to an observation space of size $1429$ and an action space with $59$ line actions (discrete actions), $66,918$ topology actions (discrete actions) and $80$ redispatch actions (continuous actions). The key challenge of the \gls{l2rpn} robustness track, however, was an adversarial agent described in \cite{omnes2021adversarial}. The adversarial agent disconnected lines in the grid quasi-randomly, simulating unforeseeable moments in the grid. In the challenge, the number of lines has been limited to a total of ten target lines. 
The robustness environment further includes specific constraints to ensure a realistic setting of the challenge \cite{marot2020learning}. One constrain was that there were several reasons for a power line to disconnect from a substation. One could be a power overflow, i.e., $\rho_{max,t}>1.0$, which would lead to a disconnection after three consecutive timesteps. Another were external line outages, either planed in the case of maintenance, or unplanned through adversarial attacks or other failures. Further, there were also constrains with regard to the action of the agent. Agents could not repeat an action on the same target (line, substation, redispatch) and therefore had to wait for a specific cooldown period. Again, as outlined in Section~\ref{sec:rel_work}, there were differences between the cooldown periods of forceful line disconnection and active line disconnection by the agent, leading to imbalance in the game mechanics.\footnote{
In a real life scenario the power lines require inspections before reconnection, thus the forcefull line disconnection had a cooldown of at least ten timesteps, while the line disconnection of the agent only had three. One timestep corresponds to five minutes in the simulation.} The agents had to adapt to the adversarial agent and these constrains, given that a missing line could trigger a cascading failure elsewhere in the grid \cite{marot2020learning}.
In terms of the evaluation metric, we decided to use the same score as the \gls{l2rpn} challenge, described in \cite{marot2021learning}. The score ranges from $[-100,0,80,100]$, where $-100$ corresponds to an initial failure, $0$ corresponds to the survival of the Do-Nothing Agent and $80$ indicates a completion of all scenarios. The maximum score of $100$ could be achieved, if the agents were further able to optimise the economical cost of their actions. Consequently, a higher score usually indicates a longer survival in the scenarios. For the testing, we were able to receive the test scenarios of the original challenge, thanks to the courtesy of the \gls{l2rpn} organisers. 

\section{The Teacher-Tutor-Junior-Senior Framework}
\label{sec:method}
\subsection{Solution by Binbinchen}
\label{sec:agent}
With the Grid2Op configuration outlined above, we analysed Binbinchen's solution \cite{binbinchen}. They proposed a \textit{Teacher-Tutor-Junior-Senior} framework in which a \gls{rl} agent was trained through imitation learning of a rule-based agent. In the first step, the \textit{Teacher} method used a brute force approach to gain experience and select topological actions by simulating all $66,918$ actions. The most frequently used actions of the \textit{Teacher} were then selected (208 by Binbinchen). With the reduced set of actions, the rule-based \textit{Tutor} gathered experience consisting of observation-action pairs. The pairs were afterwards used as ground truth in a supervised learning context for imitation learning. The pairs were fed into the \textit{Junior} model, which was a feed-forward network.
In the last step, the \gls{rl} approach, called \textit{Senior}, was trained on an adapted Grid2Op environment. The \textit{Senior} itself was started with the weights of the Junior model to accelerate faster training times and convergence.  

We decided to analyse this approach in depth for the following reasons:
First, in this paper we focus on unitary topology actions, i.e., a single topology action that changes the grid, rather than multiple consecutive actions.
This is because we want the agent to identify importance based on its own strategy, rather than giving it a pre-selected order. According to \cite{marot2021learning}, both the first place agent \cite{zhou2021action} and the third place agent \cite{lujixiang} used more sequential approaches. Further, \cite{zhou2021action} combined their actions with redispatching actions.  
In this work, we only allow topology actions and line actions in order to reconnect disconnected lines, thus favouring the Binbinchen approach. As a second reason, we chose the framework because of its clear structure and comprehensibility. In the case of \cite{zhou2021action}, interpretation is rather difficult, as the underlying method is based on a black-box optimisation approach. This is supported by the fact that Binbinchen's code is fully available, which allows for in-depth analysis and replicability. Finally, another participant reached second place with an adapted version of the Binbinchen agent in the 2021 \gls{l2rpn} Trust challenge \cite{marot2022learning}. Therefore, the \textit{Teacher-Tutor-Junior-Senior} framework is explained in detail in the following.

\subsection{The Teacher}
\label{ssec:teacher}
Although the robustness path of Grid2Op has a total of $66,918$ topological actions, not all actions are feasible. Looking at Figure ~\ref{fig:grid}, one can see that substation 16 is connected to several other substations, resulting in many possible combinations. In fact, more than 97.9\% of all topology actions are topology configurations at substation 16. It is therefore clear that there is a need to reduce the number of actions. Thus, the \textit{Teacher} model searches through all available actions in a brute force manner and selects the best candidate. More precisely, the \textit{Teacher} iterates through different scenarios of the environment until the threshold of $\rho_{teacher} = 0.925$ is exceeded by $\rho_{max,t}$.\footnote{Note that  activation thresholds differ between the \textit{Teacher, Tutor} and \textit{Senior}. These $\rho_{agent}$ were selected in preliminary experiments.} In that case, the \textit{Teacher} searches through all actions and evaluates them with the pre-build simulation method of the Grid2Op environment. The action which results in the lowest $\hat{\rho}_{max,t+1}$ is selected, executed and saved (excluding do-nothing actions). After simulating a large number of scenarios, all the saved actions are evaluated. The actions are sorted by their frequency, i.e., the number of occurrences across the scenarios. Afterwards, one manually selects a subset of the most frequent actions. 
Next to the simple brute force approach, it is possible to adjust the searching algorithm to gather more advanced actions. Binbinchen created an adversarial \textit{Teacher} that forcefully disconnected relevant lines, simulating more frequently the behaviour 
of the adversarial agent. This resulted in two different action sets of Binbinchen with 62 actions from the adversarial \textit{Teacher} and 146 of the normal \textit{Teacher}. In our work, we also use these 208 actions and build our enhancements on top. 

\subsection{The Tutor}
\label{ssec:tutor}
With the subset of the \textit{Teacher}, a rule-based agent is created to generate experience for the following stages of the framework. The respective \textit{Tutor} is already a fully functioning rule-based agent that iterates over various scenarios of the Grid2Op environment and saves both the observations, as well as the selected actions as experience.\footnote{Note that the observation of the  \textit{Tutor} were only a subset of all available observations, see \ref{app:A} for more information.} In order to achieve realistic behaviour, several rules are implemented for the \textit{Tutor}. First, the agent does not interact with the grid, if $\rho_{max,t}$ is below the \textit{Tutor} threshold of $\rho_{tutor}=0.9$. Second, if the threshold is breached, the \textit{Tutor} iterates over all 208 actions with a greedy approach. For each action, the \textit{Tutor} checks whether the action is valid and then selects the action which results in the lowest simulated $\hat{\rho}_{max,t+1}$. Third, if a line is disconnected and there is no cooldown time remaining, the agent automatically reconnects the respective line. 
With this rule-based approach, Binbinchen were able to achieve a score of $44.69$ on the online data set, which is already quite satisfactory, when comparing it to their \gls{rl} approach with a score of $52.42$. In our work, we build on top of the general \textit{Tutor} structure and developed three additional enhancements that will be discussed in Section~\ref{sec:improvements}. 

\subsection{The Junior}
\label{ssec:junior}
The third component of the framework is the \textit{Junior} agent, which is a feed-forward network that imitates the behaviour of the \textit{Tutor}. The design of the \textit{Junior} is fairly simple: The input layer of the network has the shape of the \textit{Tutor} observation, while the output layer has the shape of all 208 topological actions. In the original agent from Binbinchen, the network consists of four layers à 1000 neurons (Relu-Activation) to process the data. In our work, we executed a hyperparameter search, which resulted in a different number of neurons, as seen in \ref{app:B}.\\
With regard to the performance, the \textit{Junior} is able to correctly predict the right \textit{Tutor} action with around 37\% accuracy, which is relative low. However, when considering the top 20 actions, the correct action is found with an accuracy of 92\%. After the network is trained, the \textit{Junior} is used to jump-start the \textit{Senior} model.

\subsection{The Senior}
\label{ssec:senior}
Finally, the \gls{rl} \textit{Senior} agent has a similar neural network architecture to the \textit{Junior} model, i.e., the same layer and neuron structure as well as input and output shapes. The model is trained with the state-of-the-art \gls{rl} algorithm \gls{ppo} \cite{schulman2017proximal}. At the  beginning of the training, the model is first initialised with the weights of the \textit{Junior} model. Because the agent is only required to act in critical situations, the training of the \textit{Senior} had to be adjusted. 
No action is taken if $\rho_{max,t}$ is below the threshold of $\rho_{senior}=0.9$, with the exception of line reconnection if applicable. Is the threshold breached, the current state is passed to the \textit{Senior} and the model has to chose a valid action. In the training, the \textit{Senior} collects the basic Grid2Op reward, however accumulated over all previous steps where the do-nothing action was selected. \\
After convergence, the \gls{rl} model is combined with the heuristic strategies (do-nothing and line reconnection actions) to create the final agent. Again, an action of the \gls{rl} model is only required, if the threshold is breached. In this case, the model returns the probabilities of the \gls{rl} policies and sorts the list of actions. The list is then reviewed one by one until a suitable candidate is found. Overall, one could therefore consider the final agent as a hybrid between the rule-based approach and the \gls{rl} model.

\section{Methodological Improvements of the existing agent}
\label{sec:improvements}
As outlined in the research contributions in Section~\ref{ssec:res_contr}, we propose multiple enhancements of the \textit{Teacher-Tutor-Junior-Senior} framework of Binbinchen \cite{binbinchen}, with a primary focus on the \textit{Tutor} agent. Based on their code, we revised and enhanced the framework and created the CurriculumAgent package.\footnote{CurriculumAgent: \url{https://github.com/FraunhoferIEE/CurriculumAgent}  (last access 20/03/2023).}
The enhancements were threefold and are presented in the following. 

\subsection{N-1 Strategy Improvements}
\label{ssec:n1}
As a first improvement, we propose to prioritise topology actions that especially reduce the overall $\rho_{max,t}$ in the event of a line failure, e.g., through a lightning stroke or adversarial attacks on lines. This corresponds to the well-known principle of N-1 security, already established among \gls{tso} operators. 
We search for these topological actions by creating a special N-1 \textit{Teacher} as well as an N-1 \textit{Tutor} that prioritises the execution of the N-1 actions.\\
The underlying pseudo-code of the N-1 algorithm is given in Algorithm~\ref{alg:n1} and can be described as followed: 
We start the N-1 search for a subset of lines of size $M$ and all topological actions $a^{(topo)}_{i} \in \ \mathcal{A}^{(topo)}$ with $i= 1,\ldots,N$.
For each available topological action $a^{(topo)}_{i}$, we iterate over each line l in the subset $\mathcal S \subseteq \mathcal L$, with $l=1, \ldots, M$ and create an action $a^{(line)}_{l} \in \ \mathcal{A}^{(line)}$ that disconnects the line. By combining both actions $a_{i,l} = a^{(topo)}_{i}\ \land \ a^{(line)}_{l}$, we can then simulate the observation of the next step. With the simulated line capacities $\hat{\rho}^{(i,l)}_{j,t+1} \in \hat{\mathcal{P}}$ for all lines $j=1,\ldots,L$, we have the combined expected effect of the topology action $a^{(topo)}_{i}$ and the disconnection of line $l$. Consequently, we can then calculate the maximum value of the grid $\hat{\rho}^{(i,l)}_{max,t+1}$.\footnote{Note that while we only disconnect a subset of lines, the action itself might have an effect on other lines as well. Thus, we check the $\hat{\rho}^{(i,l)}$ for the whole grid.} \\
Afterwards, we record the maximum value across all lines $\hat{\rho}^{(i,max)}_{max}$, which are the worst possible line overloads for the topological action $ a^{(topo)}_{i}$, when one of the lines is disconnected. Consequently, by sorting all $\hat{\rho}^{(i,max)}_{max}$ in ascending order, it is possible to get the best N-1 action that is available in the respective observation, i.e. $a_i^{\star} = \argmin  (\hat{\rho}^{(i,max)}_{max})$. Note, however, that the N-1 calculation is computationally intensive. If expert knowledge is available it can be beneficial to only select a subset of the lines. 
\begin{algorithm}
\setstretch{1.35}
\begin{algorithmic}
\FORALL{$i= 1,\ldots,N$} 
    \STATE{\textrm{Select} $a^{(topo)}_{i}$}
    \FORALL{$l =  1,\ldots,M $} 
        \STATE {$a^{(line)}_{l} \gets \textrm{disconnect line } l$ }
        \STATE {$a_{i,l} \gets a^{(topo)}_{i}\ \land \ a^{(line)}_{l}$}
        \STATE {$\hat{\rho}^{(i,l)}_{j,t+1} \gets \textrm{simulate}(a_{i,l})$ }
        \STATE {$\hat{\rho}^{(i,l)}_{max,t+1} \gets \max\limits_{j=1,\ldots,L}(\hat{\rho}^{(i,l)}_{j,t+1})$ }
    \ENDFOR
    \STATE {$\hat{\rho}^{(i,max)}_{max} = \max\limits_{l=1,\ldots,M}(\rho^{(i,l)}_{max,t+1})$}
\ENDFOR 
\STATE {$i^* \gets \argmin \limits_{i=1,\ldots,N}(\hat{\rho}^{(i,max)}_{max} )$}
\RETURN $a^{(topo)}_{i^*}$
\end{algorithmic}
\caption{\\ The N-1 algorithm in pseudo code:}
\label{alg:n1}
\end{algorithm}

\subsection{Topology Reversion Improvement}
\label{ssec:topo} 
As a second improvement, the topology reversion proved to be another essential component in developing a more advanced greedy agent. This improvement is based on the idea that the electricity grid is most stable in its original state. In our case, this translates to switching to bus level one for all substations. Therefore, it is beneficial to return to the original state of the grid. \\
The improvement is implemented as follows. When the maximum line capacity falls below the reversion threshold $\rho_{rev}=0.8$, meaning no imminent danger is present, the greedy agent automatically searches through all substations and checks, whether their topology had been changed.\footnote{The $\rho_{rev}$ is explicitly lower than the $\rho_{tutor}$ because we want to make sure that the grid is stable and no agent action is required.} The agent then compares both the continuation of the current topology and a reversion to the original state in a simulation and chooses the better candidate. If multiple options are possible, the reversion with the lowest $\hat{\rho}_{max,t+1}$ is selected. 

\subsection{Code Improvements} 
\label{ssec:code_improv} 
The last enhancements were code improvements to make the Binbinchen approach usable for the \gls{l2rpn}-Baselines.\footnote{L2RPN Baselines: \url{https://l2rpn-baselines.readthedocs.io/en/latest/} (last access 20/03/2023).}
This included the requirement to make the agent compatible for all Grid2Op environments. Thus, hard coded lines and environment-specific solutions had to be reworked.
In addition to the major changes, we added smaller features, such as filters of the observation array and the ability to add scalers from the Scikit-learn package for normalisation.\footnote{Scikit-learn package: \url{https://scikit-learn.org/stable/} (last access 20/03/2023).}
Furthermore, RLlib \cite{rllib} was added to ensure a more flexible training with different \gls{rl} algorithms. Supplementary to the RLlib enhancement, we also equipped both neural networks with hyperparameter tuning, which in case of the \textit{Junior} model was the \gls{bohb} algorithm \cite{falkner2018bohb} and in case of the \textit{Senior} the \gls{pbt} algorithm \cite{jaderberg2017population}. Finally, we provided a test suite and ensured thorough documentation by adding docstrings, typehints, a readme and examples to our code to make the usability easier

\section{Research Design}
\label{sec:framework}
While creating the evaluation of the experiments, we could observe that the performance within the test scenarios varied significantly and was strongly influenced by the selected seed of the environment. Therefore, we did not evaluate the agents on one specific seed but instead chose a more robust research design. We randomly generated thirty seeds, which in turn were fed into the evaluation procedure.\footnote{The choice of a total of 30 seeds was a compromise between a sufficiently high degree of freedom and the computational cost of the evaluation. For the evaluation we use the internal method \verb|ScoreL2RPN2020| of Grid2Op.} This setting ensured that the performance can be interpreted more reliably and we recommend this procedure for other researchers as well. 
With the proposed improvements of Section~\ref{sec:improvements}, we compare a total of six agents, which are described as follows: 
\begin{enumerate}
    \item To gain a baseline result, we provide the \gls{d_n} as well as the \gls{exp} of \cite{marot2018expert} introduced in Section \ref{sec:rel_work}.
    \item Furthermore, we include \gls{t_o} of the Binbinchen solution to our analysis.
    \item Regarding the enhancements, we propose a 
    \gls{t_n1} and a \gls{t_n1_topo}.
    \item Lastly, we also provided a \gls{senior}, based on the \gls{t_n1_topo}. This \gls{senior} is a hybrid between the \gls{rl} model and heuristic methods (line reconnection, topology reversion). 
\end{enumerate}
Note that we only included the \gls{t_o} of the Binbinchen solution for two reasons. First, we enhanced primarily the \textit{Tutor} agent, thus comparing the tutors is most interesting.
Second, on the Grid2Op version 1.7.1, the  Binbinchen \textit{Senior} did perform poorly on all thirty seeds. \\
For both the \gls{t_n1} and \gls{t_n1_topo} improvements, as well as the \gls{senior} it was necessary to compute the N-1 search, which was done by an N-1 \textit{Teacher}. The resulting 300 actions were added to the base action set of 208 actions. In terms of the \gls{senior}, the agent was trained with all 508 actions (208 and 300 N-1) on the RLlib \cite{rllib} framework. A hyperparameter training was included for both \textit{Junior} and \textit{Senior} models, as described in Section \ref{ssec:code_improv}. The hyperparameter can be found in \ref{app:B}. To prevent cherry picking, we run three experiments in parallel and evaluate the checkpoints in a validation environment, based on the training scenarios. We then select the best performing checkpoint across all three experiments.
\begin{table*}[!h]
\parbox{.6\linewidth}{
\caption{Summary of the agents' results. All agents were run on the robustness track of the 2020 \gls{l2rpn} test environment(24 scenarios) with thirty different seeds. The performance across the seeds is recorded below. We list the mean and the standard deviation in the first column and the median as well as the 25\% and 75\% quantile in the second column. Note that the \gls{d_n} agent achieves a score of $0.00$ per default.}
\label{tab:result_seed_short}
\begin{tabular}{l|l|l}
\toprule
 Agent & Mean (Sd)& Median (Q25, Q75) \\
\midrule
\gls{d_n} & 00.00 (0.00) & 00.00 (00.00, 00.00) \\
\gls{exp} & 26.45 (4.52) & 26.69 (24.03, 29.88) \\
\gls{t_o} & 38.44 (4.19)& 37.89 (35.75, 40.81) \\
\gls{t_n1} & 41.88 (4.11)& 40.80 (38.97, 45.75) \\
\gls{t_n1_topo} & 48.90 (4.67)& 48.39 (44.67,53.40) \\
\gls{senior} & \textbf{49.12} (4.08)& \textbf{48.70} (45.53, 51.15) \\
\bottomrule
\end{tabular}
}
\hfill
\parbox{.38\linewidth}{
\caption{Test Results of the Welch's t-test \cite{welch1947generalization} with the hypothesis $H_0: \mu_i = \mu_j$  against the alternative hypothesis $H_1:\mu_i \neq \mu_j$. For the normality assumption we tested with  D'Agostino test \cite{d1973tests} and could not reject the $H_0$ hypothesis, so non-normality could not be suspected.}
\label{tab:t_test}
\begin{tabular}{l|l}
\toprule 
$H_0$ Hypothesis & p-value\\
\midrule 
$H_0: \mu_{T_{o}} = \mu_{T_{N-1}}$ & 0.0022 \\
$H_0: \mu_{T_{o}} =  \mu_{T_{N-1 Topo}}$ & 8.7e-13\\
$H_0:  \mu_{T_{N-1}}= \mu_{T_{N-1 Topo}}$ & 6.9e-08\\
$H_0:  \mu_{S}= \mu_{T_{N-1 Topo}}$ & 0.8454\\
\bottomrule
\end{tabular}
}
\end{table*}
\section{Results}
\label{sec:results}
\begin{figure*}[ht]
    \centering
    \caption{Visualisation of the average survival time from the \gls{senior} (blue), the \gls{t_n1_topo} (red), the \gls{t_n1} (green), the \gls{t_o} (purple), the \gls{exp} (orange) and the \gls{d_n} (turquoise). Each bar represents the average survival time of the agent in the respective scenario across all seeds. The overall average is reported in the legend.}
    \includegraphics[trim={0cm 0.2cm 0cm 0.0cm},clip, width=1.0\linewidth]{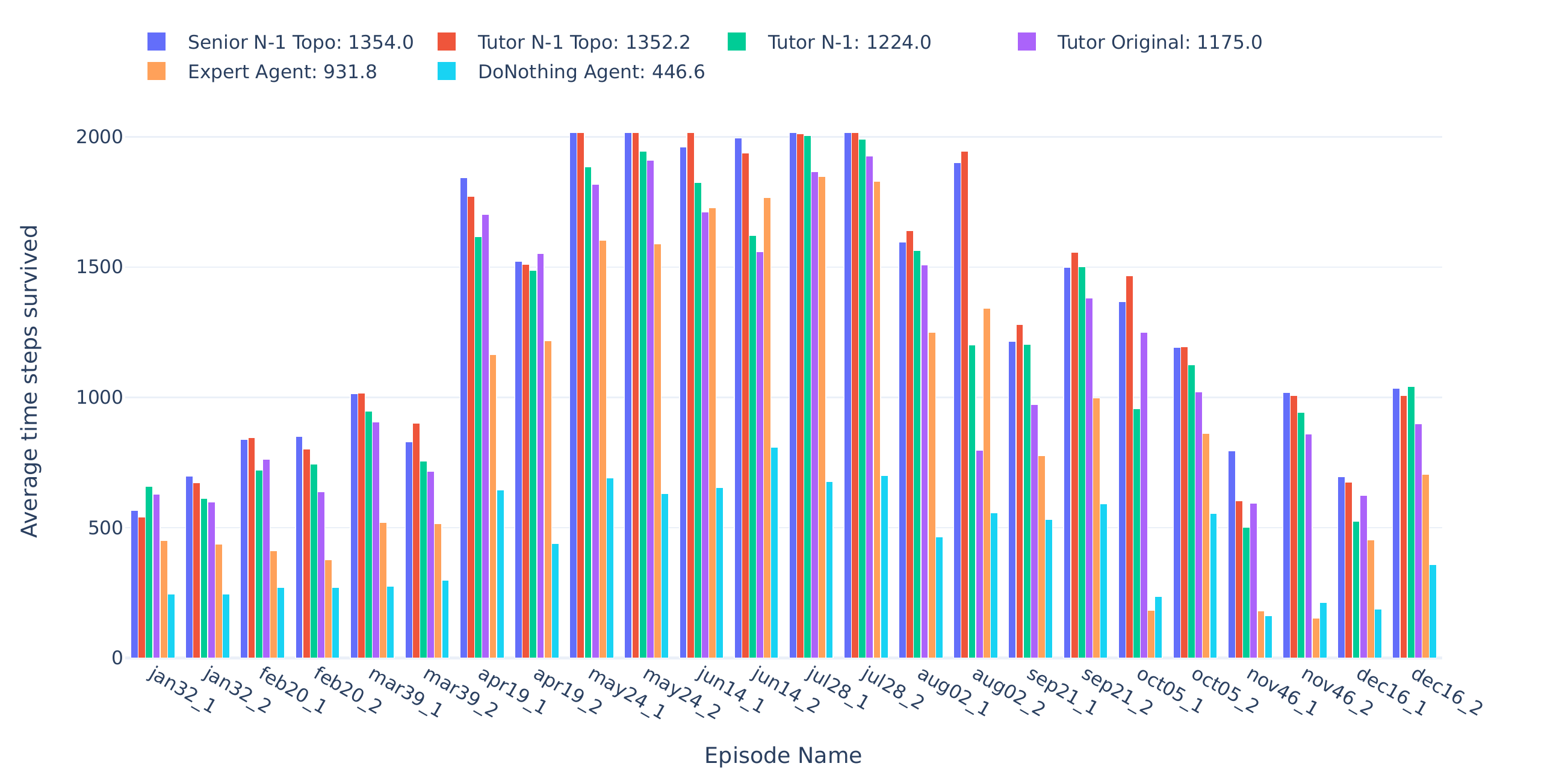}
    \label{fig:survival}
\end{figure*}
\subsection{Experiment Results}
\label{ssec:performance}
Using the experimental framework outlined above, the agents were evaluated on the thirty random seeds. The results can be found in Table~\ref{tab:result_seed_short}, where we summarise the scores across seeds. A larger table and a Boxplot can be found in the \ref{app:C}.
In terms of our research question, we are primarily interested in the performance of the rule-based agents. Here, we could observe a clear improvement. While the \gls{t_o} reached a mean score of $38.44$, the \gls{t_n1} reached $41.88$ and the \gls{t_n1_topo} even $48.90$, which is an increase by $27\%$. We tested with the Welch's t-test \cite{welch1947generalization}, whether the results of the different \textit{Tutors} are from the same distribution and were able to reject the $H_0$ hypothesis for all Tutors, see Table \ref{tab:t_test}. This shows that both the N-1 strategy and the reversion to the topology significantly improved the score of the agents. Further,  the mean and median are relatively similar, thus we cannot directly detect a high influence of outliers. However, on can see in the quantiles that there is indeed a large variation in the performance depending on the seed. \\

Next to the \textit{Tutors}, we also included the \gls{senior} in our experiments to evaluate the impact of the \gls{rl} algorithms. In this regard, we observe that the agent was slightly better than the \gls{t_n1_topo} with a score of $49.12$. However, the improvement was only fractional and not statistically significant, as shown in Table \ref{tab:t_test}. Further note that the \gls{senior} had a better performance in the median and the 25\% quantile, but the \gls{t_n1_topo} with $53.40$ a better 75\% quantile. This is incredibly interesting, given that researchers often postulate a clear superiority of \gls{rl} approaches, which we could not replicate. Instead, it seems that the success is rather highly depending on the correct action sets. With respect to the baseline agents, the \gls{d_n} and \gls{exp} did not reach the score of the other agents, which is not surprising. \\ 
After evaluating the scores, we are also interested in the survival times of the different approaches. In Figure~\ref{fig:survival} we show the average survival times across the 30 seeds, divided into the 24 test scenarios. Note that all scenarios ended after 2006 steps. If an agent was able to reach an average of 2006 steps for a scenario, this means that it was able to complete the scenario in every run. 
One can see that the performance across the scenarios differed between the agents, but there were also some similarities. The winter scenarios from November up until March were harder to solve by all participants, while the summer scenarios were easier to complete. In terms of survival, only a handful were completely survived by some of the agents. Both \gls{senior} and \gls{t_n1_topo} were able to complete four scenarios in all seeds, with \gls{senior} completing the scenarios $(may24_1,may24_2,jul28_1,jul28_2)$ and \gls{t_n1_topo} completing the scenarios $(may24_1,may24_2,jun14_1,jul28_2)$. All other agents failed to achieve this task. Nevertheless, \gls{t_n1} was almost able to complete $jul28_1$. This shows a clear performance increase.\\

\subsection{Detailed Behaviour analysis of the agents}
\label{ssec:seed}
\begin{figure*}[t]
\centering
\caption{Visualisation of the different topological actions of the agents \gls{t_o}, \gls{t_n1_topo} and \gls{senior}. The actions are sorted according to the frequency of their substations. The most frequently used substation is at the top right, then all the other substations are ordered counterclockwise.  The colour coding is the same for all three agents.}
\begin{subfigure}{.3\textwidth}
  \centering
  \includegraphics[trim={4.5cm 2.1cm  4.5cm 1.6cm},clip, width=1.0\linewidth]{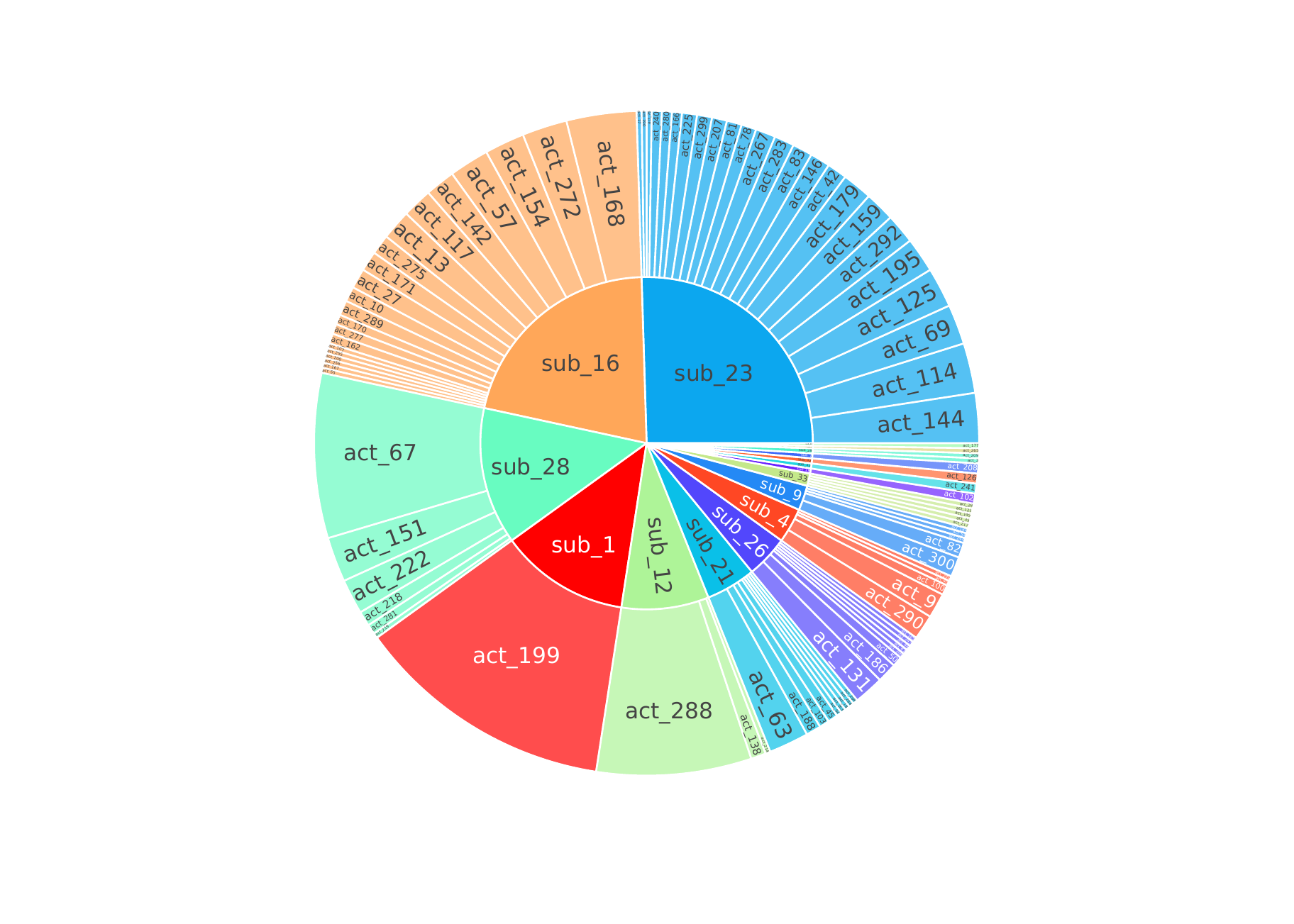}
  \caption{Actions of \gls{t_o}}
  \label{fig:sub1}
\end{subfigure}  \ 
\begin{subfigure}{.3\textwidth}
  \centering
  \includegraphics[trim={4.5cm 2.1cm  4.5cm 1.6cm},clip, width=1.0\linewidth]{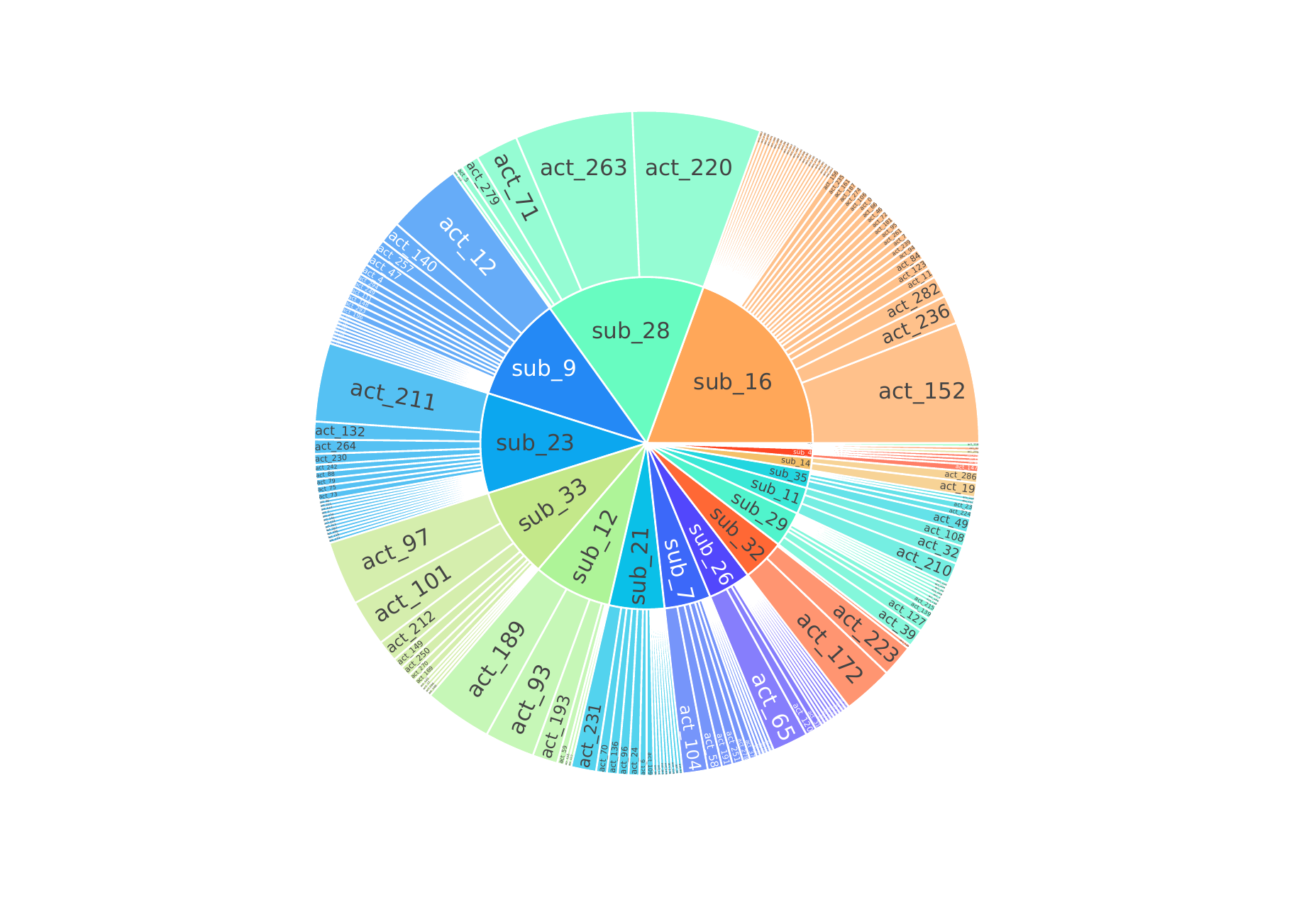}
  \caption{Actions of \gls{t_n1_topo}}
  \label{fig:sub2}
\end{subfigure} \ 
\begin{subfigure}{.3\textwidth}
  \centering
  \includegraphics[trim={4.5cm 2.1cm  4.5cm 1.6cm},clip, width=1.0\linewidth]{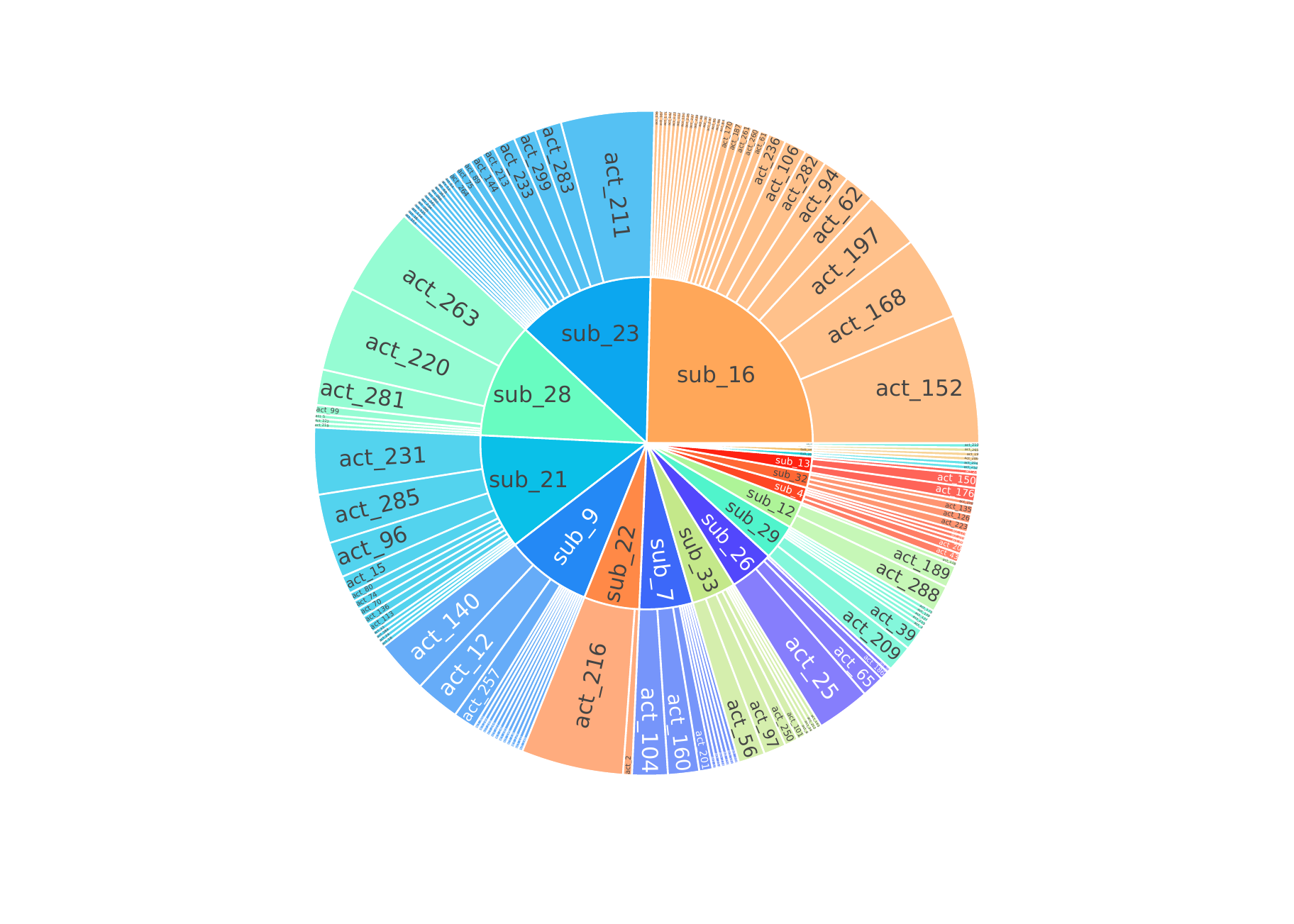}
  \caption{Actions of \gls{senior}}
  \label{fig:sub3}
\end{subfigure}
\label{fig:substations}
\end{figure*}
After analysing the overall performance across the thirty seeds, we further provide a more detailed evaluation regarding the behaviour of the agents. Thus, we randomly select a run and looked at various metrics available in Grid2Op.\footnote{The randomly selected seed had a similar performance to the average score of Section \ref{ssec:performance}. Thus, we expect the behaviour to be comparable. Note that the data extraction is based on adjusted methods of the package Grid2Bench \url{https://github.com/IRT-SystemX/Grid2Bench} (last access 20/03/2023).}
To begin with, we are interested in individual topological behaviour on a substation level. Hence, in Figure~\ref{fig:substations} we show the topological actions sorted by their specific substation. First of all, we can see that the \gls{t_o} executed actions from only five substations in more than 80\% of the time. In contrast, the distributions between the substations of the \gls{t_n1_topo} and the \gls{senior} were more diversified with the 300 N-1 actions. Based on the grid of Figure~\ref{fig:grid}, we see that all three agents centred their actions around the three major nodes 16, 23 and 28, which were necessary to keep the grids stable. However, while the substation 23 took about one forth of the overall actions from the \gls{t_o}, the N-1 agents rather preferred topological changes on the substation 16. It is also interesting to note that the \gls{t_o} only had one action at substation one, which it used frequently. The N-1 agents, on the other hand, did not rely on this substation at all. \\
When comparing the behaviour between the rule based \gls{t_n1_topo} and the \gls{senior}, we can see that there are several similarities but also slight changes in the behaviour. One can observe that the rule-based agent utilised the substations 33 and 9 to divert power from substation 16 to the south-east and north-west. In contrast, we see that these substations rank lower for the \gls{rl} agent and instead the \gls{senior} diverted the power to the substations 23 and 21, i.e., corresponding to a north-eastern route. As a result, the \gls{senior} had to rely on substation 22 more often. This shows a different learned strategy by the \gls{rl} approach, with the same potential outcome.\\
\begin{figure}[ht]
\begin{minipage}{.48\textwidth}
    \centering
    \caption{Display of the computation time for each action. The left graph shows the computation time, where each point is one action of the agent. The vertical lines correspond to a specific scenario. On the right, we aggregate the computation time across all scenarios in a rug plot. Note that for comparison, we only include the computation times if all three agents survived until the given time step.}
    \includegraphics[trim={0cm 0.5cm 0cm 0cm},clip, width=1.1\linewidth]{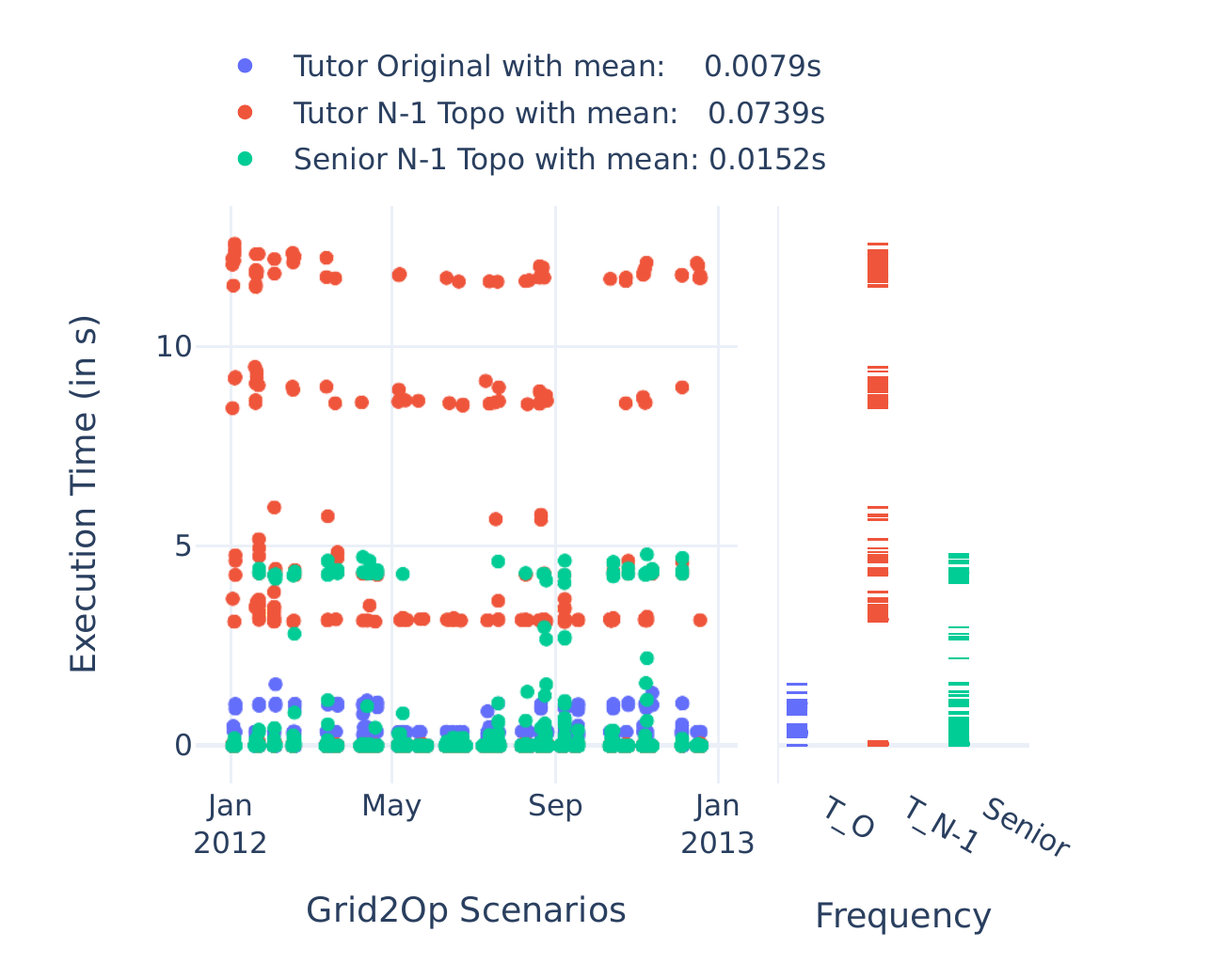}
    \label{fig:computation_time}
\end{minipage}
\hfill
\begin{minipage}{.48\textwidth}
    \centering
    \caption{Distance in Topology for the first July scenario. The y-axis describes the number of substations that differ in comparison to the original topology. Note that all agents completed the scenario.}
    \includegraphics[trim={0cm 1.4cm 1cm 1.5cm},clip, width=1.0\linewidth]{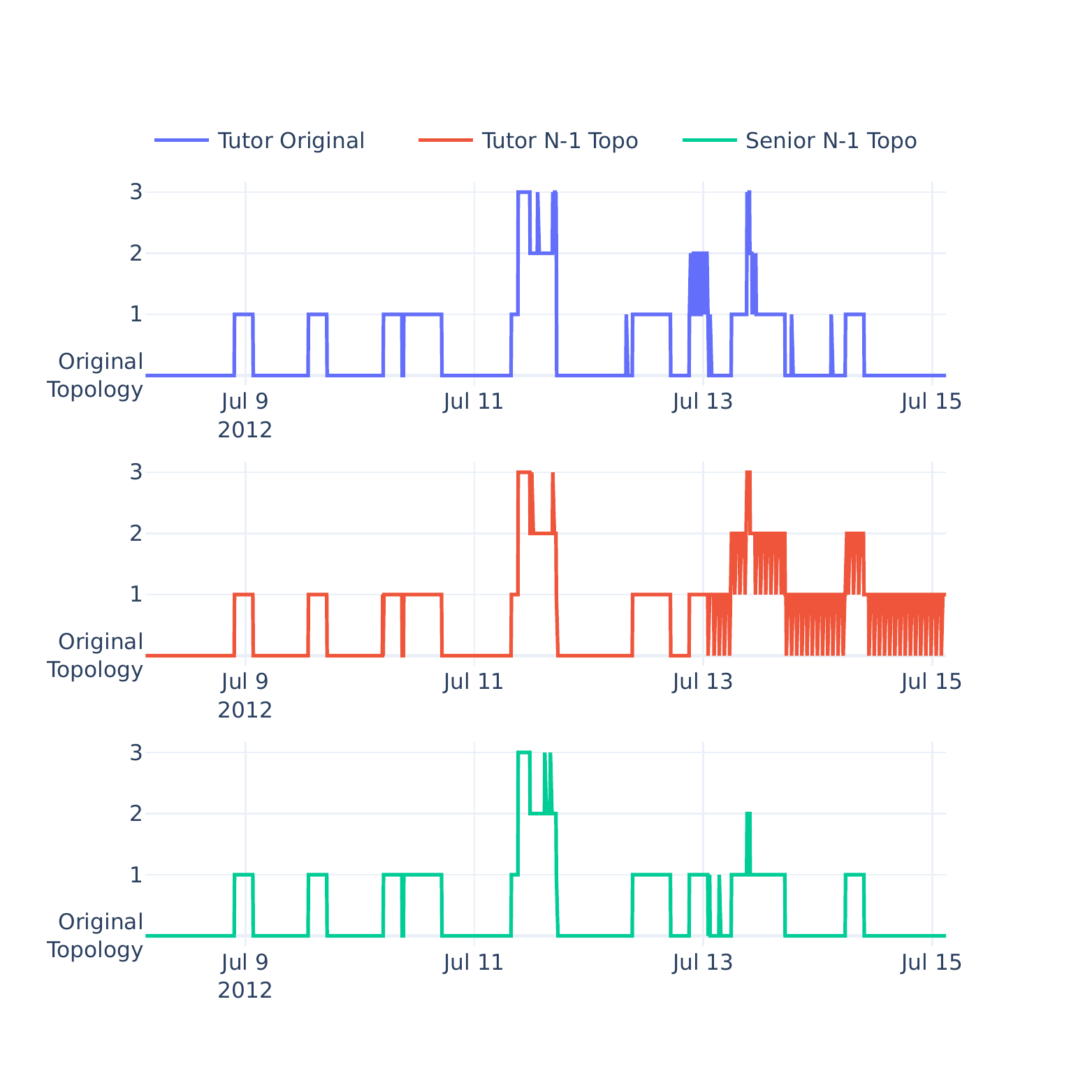}
    \label{fig:topo_distance}
\end{minipage}
\end{figure}
Next to the different routing, we want to highlight different computational times of the agents. In Figure~\ref{fig:computation_time}, we plotted for each step the required computation time of the agent, which we summarised in the rug plot on the right side.
The first thing that becomes clear is that the \gls{t_n1_topo} increased the computation time quite a bit compared to the \gls{t_o}. This can be explained by the fact that the \gls{t_o} had the 208 actions divided into two sets of size 62 and 146 and first only evaluated the 62 actions. Only if no adequate candidate was found, it continued with the 146 action set. Further, no additional heuristics were added to the tutor. In contrast, the \gls{t_n1_topo} started with the N-1 action set consisting of 300 actions, then continued with 62 and the 146 actions. Consequently, the longer survival of \gls{t_n1_topo} came at a cost in terms of computation time.\\
Compared to the \gls{t_n1_topo}, the \gls{senior} was faster than the rule-based approach. This is due to the fact that the \gls{senior} did not have a clear division of its action sets, as it provided a policy for all 508 (208+300) actions. This made it easier to find a suitable candidate and the search was accelerated without any loss of accuracy. Note that the actions with a computational time of almost zero correspond to the do-nothing actions, which all three agents selected when the grid was stable.
As a last result, we also want to look at the distance from the original topology, which we visualised exemplary for the July week $jul28\_1$ in Figure~\ref{fig:topo_distance}. The distance was measured in the number of changed substations in comparison to the original topology configuration. Here, there are two interesting things to note. First, we can see that all agents had a relatively similar behaviour until the 12th of July, moving up to three steps away from the original topology and afterwards returning to it. However, we see a clear difference between the \textit{Tutors} and the \gls{senior} at the start of July 13th. While the \gls{senior} had only a maximum distance of two topology changes, the \textit{Tutors} had a distance of three, thus showing that the \gls{senior} found a better action in that moment. A second thing to note is that the \gls{t_n1_topo} showed very unstable behaviour after 13 July, which is not ideal for a real substation. A possible explanation could be that the $\rho_{max,t}$ of the grid was fluctuating around the threshold $\rho_{tutor}$ of the \gls{t_n1_topo}, thus triggering the need to act. Both results are very interesting, because they show that the \textit{Senior} was able to achieve a similar effect, however used a different topology action. Here, less distance from the original grid was required and the overall state was more stable.  \\
\subsection{Discussion and Future Outlook}
\label{ssec:discussion}
As outlined in the previous Section~\ref{sec:results}, we were able to significantly improve the original greedy search \textit{Tutor} by providing both an active revision of the topology as well as an additional N-1 strategy. In order to obtain more advanced rule-based approaches, we showed that classical power-grid strategies can be beneficial as improvements. \\
When comparing the more sophisticated rule-based agent with the \gls{rl} approach, we were not able to demonstrate a clear superiority. Even though the \gls{rl} agent was slightly better, one could argue that this may not have enough statistical significance.
This is particularly interesting when considering that  heuristic methods were also included in the \gls{senior}. This shows that strategies such as reconnecting lines and simulating prior to the execution of an action have a high proportion of the overall performance score. 
Nevertheless, we still consider \gls{rl} to be a necessity for future network operations for the following three reasons:\\
First, the \gls{rl} approach clearly showed an advantage in the computational speed, as was shown in Figure~\ref{fig:computation_time}. This is quite interesting considering that the agent had no direct goal to use N-1 actions like the \gls{t_n1_topo}. However, the agent was still able to learn a similar behaviour, demonstrating the ability to imitate desired behaviour. This feature may be crucial, especially for larger grids. Considering that an increase in the number of available actions could push rule-based approaches to their limits, \gls{rl} can still compute solutions in an acceptable time.
Second, in order to replicate the Binbinchen approach we only used the \gls{ppo} algorithm in this work. We therefore did not focus on specific \gls{rl} improvements. As other researchers showed, e.g, in \cite{dorfer2022power}, there are advanced \gls{rl} techniques available that can enhance the \gls{rl} models. Thus, it could be possible to further increase the score through different \gls{rl} agents. Third, it can be advantageous to consider multiple approaches simultaneously for ensemble methods. As analysed in Section~\ref{ssec:seed}, the behaviour of the \gls{senior} differed from that of the \gls{t_n1_topo} agent, thus it could provide alternative strategies for grid operators.\\
As a consequence, although the main development of this article was a more sophisticated rule-based agent, we encourage other researchers to further improve \gls{rl} approaches. One possibility might be a split of the \gls{senior} into two independent agents, in order to gain more specialised agents, e.g., one N-1 agent and one agent for extreme cases $(\rho_{max} > 1.0)$.

\section{Conclusion}
\label{sec:conclusion}
In this paper, we analysed the structure of the Binbinchen agent from the 2020 \gls{l2rpn} robustness Challenge and proposed additional improvements to the rule-based greedy agent. The novelty of our improvements was on the one hand the N-1 strategy and the reversion back to the base topology of the agent. In order to evaluate the proposed improvements, we tested the agents on the \gls{l2rpn} test environment with 30 different seeds and additionally evaluated one random seed in particular. Our experiments show that the improvements increased the performance by $27\%$ and that the N-1 rule-based agent was able to achieve a similar result to the \gls{rl} agent. In our detailed analysis, we also showed that by considering the N-1 actions, the overall set of actions became more diverse, leading to an increase in stability of the grid. We  discussed these results, highlighting the comparison between the rule-based approach and the \gls{rl} agent.

\section*{Acknowledgement}
This work was supported by the Competence Centre for Cognitive Energy Systems of the Fraunhofer IEE and the research group Reinforcement Learning for cognitive energy systems (RL4CES) from the Intelligent Embedded Systems of the University Kassel. 
\bibliographystyle{elsarticle-num}      
 \bibliography{ai2go_bib}

\newpage
\appendix
\label{sec:appendix}
\onecolumn
\begin{landscape}

\section{Filter of the Tutor model}
\label{app:A}
\begin{table}[!h]
    \centering
    \caption{In order to not inflate the input of the deep learning models, the observations of the Tutor were filtered to include  only the relevant data. Consequently, the experience of the \textit{Tutor} as well as the input of the \textit{Junior} and \textit{Senior} all had the reduced observation, which in case of the \gls{l2rpn} robustness track corresponds to a size of (1221,). This process was copied from Binbinchen \cite{binbinchen}. The following information was included in the observation of the \textit{Tutor},\textit{Junior} and \textit{Senior}. The description is from the Grid2Op documentation and can be found under \url{https://grid2op.readthedocs.io/en/latest/observation.html}. }
    \begin{tabular}{l|l|l}
        Category & Variable & Description \\
        \hline
        \multirow{5}{*}{Time Values} &  $month$   & Month of the Scenario \\
        &   $day$ & Day in time step $t$ \\
        &   $hour\_of\_day$ & Hour of time step $t$ \\
        &   $minute\_of\_hour$ & Minute of time stpe $t$ \\
        &   $day\_of\_week$ & Weekday of time step $t$ \\
        \hline
       \multirow{4}{*}{Generation and Load}  &   $gen\_p,gen\_q$ & Active and reactive production value of each generator  \\
        & $gen\_v$ & Voltage magnitude of  each generator at connected bus \\
        &   $load\_p,load\_q$ & Active and reactive load value of each consumption \\
        & $gen\_v$& Voltage magnitude of  each consumption at connected bus \\
        \hline
         \multirow{9}{*}{Lines} &   $p\_or,q\_or$ & Active and reactive  power flow at the origin end of each power line \\
         &$v\_or$ & Voltage magnitude at the origin end of each power line \\
         &$a\_or$ & Current flow at the origin end of each power line \\
        &  $p\_ex,q\_ex$ & Active and reactive power flow at the extremity end of each power line \\
        & $v\_ex$ & Voltage magnitude at the extremity end of each power line\\
        & $a\_ex$ & Current flow at the extremity end of each power line\\
        &$rho$ or $\rho$ &  capacity of each power line\\
        &  $line\_status$ & Whether the line is connected or disconnected \\
        &  $timestep\_overflow$ & Number of time steps since overflow  \\
        \hline
        \multirow{1}{*}{Bus Information} & $topo\_vect$ & Vector of the topology configurations\\
        \hline
        \multirow{2}{*}{Cooldowns} & $time\_before\_cooldown\_line$ & Time before an action could be done on the line \\
        & $time\_before\_cooldown\_sub$ & Time before an action could be done on the substation \\
        \hline
        \multirow{2}{*}{Maintenance} & $time\_next\_maintenance$ & Time when the next maintenance of line is planned \\
        &$duration\_next\_maintenance$& Duration of planed maintenance \\
        \hline
    \end{tabular}
    \label{tab:filtered_obs}
\end{table}
\end{landscape}
\newpage
\section{Structure Deep Learning Models}
\label{app:B}
\subsection{Visualisation Junior Model}
\begin{figure}[!h]
    \centering
    \caption{Visualisation of the Junior Model after hyperparameter training, which included the number of neurons. In the original paper, all four hidden models had a total of 1000 layers. The hyperparameter after optimisation are listed in the Table \ref{tab:hyperparam_junior} below. The observation of the Tutor model and the output corresponds to the number of actions,i.e., 208 actions from Binbinchen and 300 N-1 actions.}
    \includegraphics[trim={5cm 3.0cm 5.0cm 3.0cm},clip, width=1.0\linewidth]{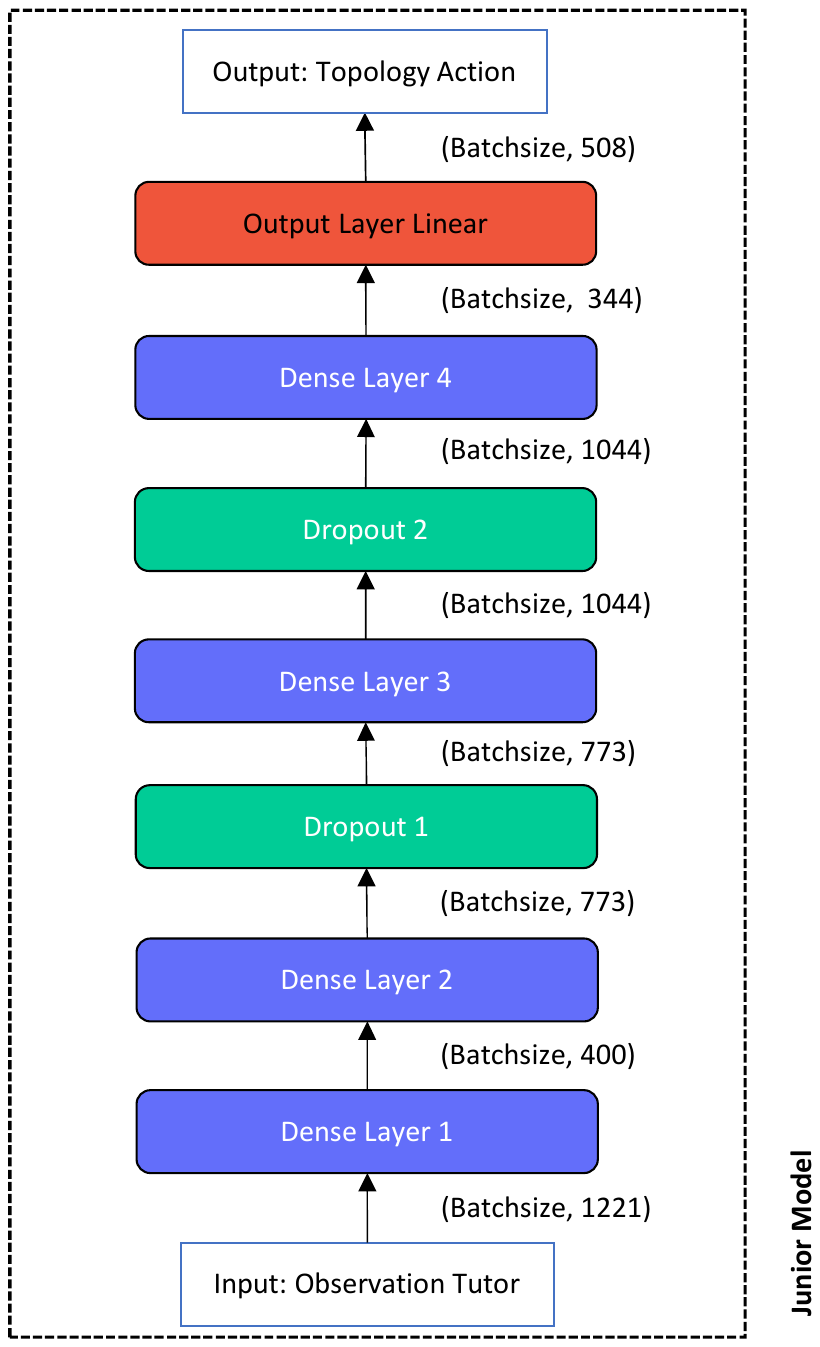}
    \label{fig:junior}
\end{figure}

\newpage
\subsection{Hyperparameter Junior Model}
\begin{table}[!h]
    \centering
    \caption{Hyperparameter of the Junior Model after running the Hyperparameter search with the \gls{bohb}\cite{falkner2018bohb}.}
    \begin{tabular}{l|l|r}
        Specification & Parameter & Value \\
        \hline
        \multirow{8}{*}{Model architecture} &        Hidden Layer 1 & 400 Neurons \\
        &   Hidden Layer 2 & 773 Neurons \\
        &   Hidden Layer 3 & 1044 Neurons \\
        &   Hidden Layer 4 & 344 Neurons \\
        &   Output Layer Linear & 508 Neurons \\
        &   Activation & Relu Activation \\
        &   Batchsize & 768 \\
        &   Dropout 1 & 0.157474 \\
        &   Dropout 2 & 0.408924 \\
        &   Initialiser & Orthogonal \\
        &   Learning Rate & 0.000128 \\
        &   Epochs & 1000 \\
        &   Early stopping & 50 steps \\
        \hline
    \end{tabular}
    \label{tab:hyperparam_junior}
\end{table}

\subsection{Hyperparameter Senior Agent}
\begin{table}[!h]
    \centering
    \caption{Hyperparameter of the Senior model. The model is trained with the \gls{ppo} algorithm in combination with \gls{pbt} \cite{jaderberg2017population}. The underlying structure of the \gls{rl} model is the same as the Junior model in Figure \ref{fig:junior}. As training framework we use RLlib \cite{rllib}.}
    \begin{tabular}{l|l|r}
        Specification & Parameter & Value \\
         \hline
        \multirow{8}{*}{\gls{ppo} Specifications} &  Learning Rate &  1e-05\\
        &   Clip-Parameter & 0.07267018 \\
        &   Entropy Coefficient & 0.00096065 \\
        &   Gamma & 0.99236394 \\
        &   VF loss coefficient & 0.84545350\\
        &   Number of SGD steps & 4 \\
        &   SGD Minibatch Size & 128(default) \\
        &   KL Coefficient & 0.2 (default) \\
        \hline
        \hline
        \multirow{8}{*}{Model architecture} &        Hidden Layer 1 & 400 Neurons \\
        &   Hidden Layer 2 & 773 Neurons \\
        &   Hidden Layer 3 & 1044 Neurons \\
        &   Hidden Layer 4 & 344 Neurons \\
        &   Activation & Relu Activation \\
        &   Initialiser & Orthogonal \\
    \end{tabular}

    \label{tab:my_label}
\end{table}

\newpage
\section{Visualisation Results}
\label{app:C}
\begin{figure}[!h]
    \centering
    \caption{Boxplot of the agent results across all 30 seeds. Each point corresponds to the total score of all scenarios for one seed. In the legend we report the median across all seeds. Note that the \gls{d_n} agent scores per default $0.00$.}
    \includegraphics[trim={0cm 0.0cm 0.0cm 0.0cm},clip, width=1.0\linewidth]{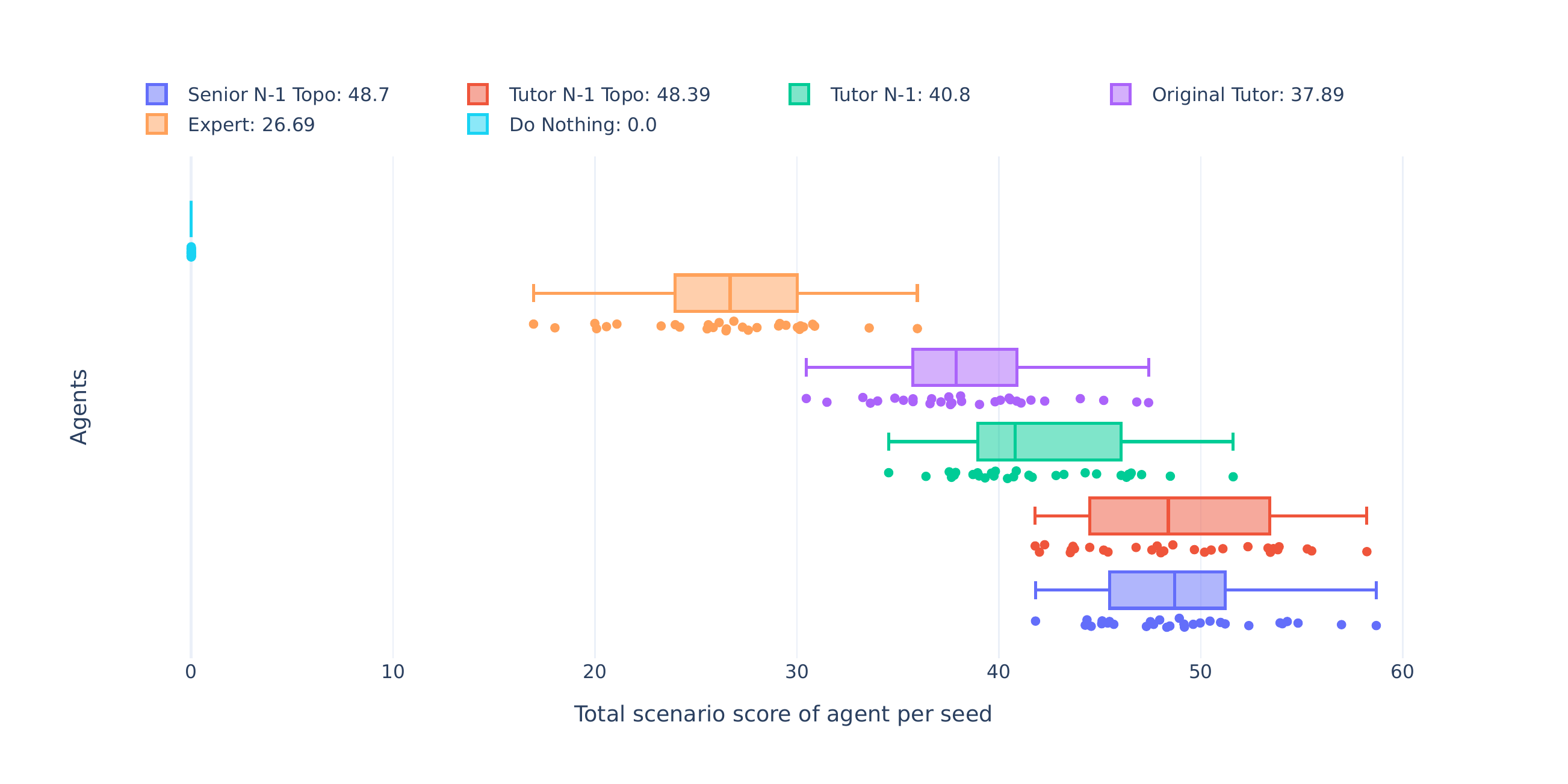}
    \label{fig:boxplot}
\end{figure}

\begin{table}[!h]
    \caption{Large summary table, similar to Table \ref{tab:result_seed_short} of the agents' results. All agents were run on the robustness track of the 2020 \gls{l2rpn} test environment(24 scenarios) with thirty different seeds. The performance across the seeds is recorded below. We list the mean, the standard deviation, the median, the 25\% and 75\% quantile  and the minimum and maxiumum values across the seeds. Note that the \gls{d_n} agent achieves a score of $0.00$ per default.}
    \label{tab:result_seed_long}
    \centering
    \begin{tabular}{lrrrrrr}
    \toprule
    & \thead{\gls{d_n}} & \thead{\gls{exp}} & \thead{\gls{t_o}} & \thead{\gls{t_n1}}& \thead{\gls{t_n1_topo}} & \thead{\gls{senior} } \\
    \midrule
    mean  &  0.0 &  26.45 & 38.44 & 41.88 & 48.90 &  49.12 \\ 
    std&  0.0 &4.52 &  4.19 &4.11 &  4.67 &4.08 \\
    Min&  0.0 &  16.95 & 30.46 &34.54 & 41.80 &  41.82 25\\ 
    \% Quantile &  0.0 &  24.03 & 35.75 &38.97 & 44.67 &  45.53 \\
    Median &  0.0 &  26.69 & 37.89 &40.80 & 48.39 &  48.70\\ 
    75\% Quantile &  0.0 &  29.88 & 40.81 &45.75 & 53.40 &  51.15 \\
    Max &  0.0 &  35.96 & 47.42 &51.61 & 58.23 &  58.69 \\
    \bottomrule
    \end{tabular}
\end{table}
\end{document}